\documentclass[final]{cvpr}
\usepackage{subcaption}
\usepackage{floatrow}
\usepackage{multirow}
\usepackage{times}
\usepackage{epsfig}
\graphicspath{ {images/} }
\usepackage{graphicx}
\usepackage{amsmath}
\usepackage{amssymb}

\newcommand\Mark[1]{\textsuperscript#1}

\usepackage[pagebackref=true,breaklinks=true,colorlinks,bookmarks=false]{hyperref}

\def\cvprPaperID{****} 
\def\confYear{CVPR 2021}

\begin{document}

\title{A Two-stage Deep Network for High Dynamic Range Image Reconstruction}

\author{S M A Sharif \Mark{1} , Rizwan Ali Naqvi \Mark{2}\thanks{Corresponding author} , Mithun Biswas \Mark{1}, Sungjun Kim \Mark{3} \\
\Mark{1} Rigel-IT, Bangladesh, \Mark{2} Sejong University, South Korea, \Mark{3} FS Solution, South Korea \\
{\tt\small \{sma.sharif.cse,mithun.bishwash.cse\}@ulab.edu.bd, rizwanali@sejong.ac.kr, sung92k@gmail.com }

}

\maketitle

\begin{abstract}

Mapping a single exposure low dynamic range (LDR) image into a high dynamic range (HDR) is considered among the most strenuous image to image translation tasks due to exposure-related missing information. This study tackles the challenges of single-shot LDR to HDR mapping by proposing a novel two-stage deep network. Notably, our proposed method aims to reconstruct an HDR image without knowing hardware information, including camera response function (CRF) and exposure settings. Therefore, we aim to perform image enhancement task like denoising, exposure correction, etc., in the first stage. Additionally, the second stage of our deep network learns tone mapping and bit-expansion from a convex set of data samples. The qualitative and quantitative comparisons demonstrate that the proposed method can outperform the existing LDR to HDR works with a marginal difference. Apart from that, we collected an LDR image dataset incorporating different camera systems. The evaluation with our collected real-world LDR images illustrates that the proposed method can reconstruct plausible HDR images without presenting any visual artefacts. Code available : \url{https://github.com/sharif-apu/twostageHDR_NTIRE21}.

\end{abstract}

\section{Introduction}

Due to numerous hardware limitations, digital cameras are susceptible to capture a limited range of luminance. Subsequently, such hardware deficiencies drive most standalone devices to capture over/under-exposed images with implausible perceptual quality \cite{liu2020single}. To counter such inevitable consequences, typically, digital camera leverage multiple LDR shoots with different exposure settings \cite{debevec2008recovering}. Regrettably, such multi-shot LDR to HDR recovery is also far from the expectation and can incorporate limitations, including producing ghost artefacts in dynamic scenes captured with hand-held cameras \cite{liu2020single, marnerides2018expandnet}. 

Contrarily, recovering HDR images from a single-shot image consider among the most prominent solution to address the shortcomings of its multi-shot counterparts. However, a single-shot HDR recovery always remains a challenging task as it aims to recover significantly higher pixel-wise information than a legacy LDR image (i.e., 8-bit image) \cite{eilertsen2017hdr}. Most notably, such LDR to HDR mapping has to incorporate dynamic bit-expansion, noise suppression, and estimation of CRF without having any additional information from the neighbour frames.

In the recent past, several methods \cite{eilertsen2017hdr, liu2020single, marnerides2018expandnet, khan2019fhdr} have attempted to reconstruct HDR images from single-shot LDR input by leveraging the convolutional neural networks (CNNs). Typically,  these deep methods learn to hallucinate the CRF and perform bit-expansion from a convex set of data samples \cite{liu2020single, eilertsen2017hdr}. Notably, the hardware-related information, explicitly the CRF  is proprietary property of the original equipment manufacturer (OEMs) and mostly remains undisclosed. Therefore, addressing the single-shot LDR to HDR mapping with a single-stage deep network with pre/post-processing operation can result in inaccurate CRF estimation along with quantization. Subsequently, such HDR mapping methods can end up with visual artefacts in real-world scenarios \cite{liu2020single}.

\begin{figure*}[!htb]
\centering
\includegraphics[width=\textwidth,keepaspectratio]{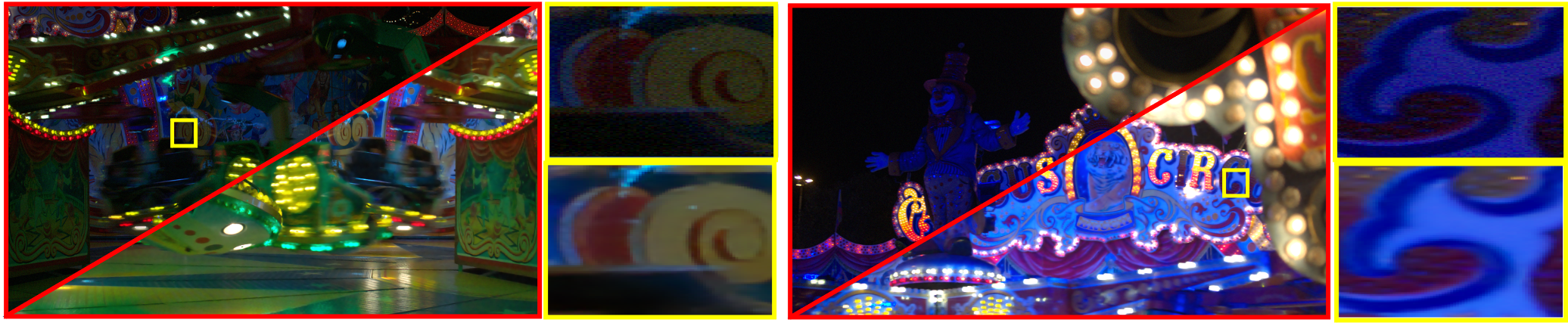}
\caption{  Single-shot LDR to HDR reconstruction obtained by the proposed two-stage deep method. The proposed network intends to map an 8-bit LDR input into a 16-bit HDR image. However, for better visualization, we normalized the reconstructed images and compared them with their inputs. In each pair, the top section illustrates the LDR input, and the bottom segment shows the corresponding HDR output. }
\label{intro}
\end{figure*}

In this paper, we propose a two-stage learning-based deep method to tackle the challenging single-shot HDR reconstruction. The proposed method comprises a two-stage deep network and learns from a convex set of single-shot 8-bit LDR images to reconstruct 16-bit HDR images comprehensively (please see Fig. \ref{intro}). Here, the first stage of the proposed method performs the basic enhancements task like exposure correction \cite{yuan2012automatic, cao2020over}, denoising \cite{buades2005review,tian2020deep}, etc., and the second stage recovers the 16-bit HDR image, including the tone mapping \cite{mantiuk2009color,ledda2005evaluation}. Notably, we encouraged our network to directly learn to reconstruct HDR images without explicitly estimating hardware-related information like CRF and bit-expansion. Hence,  our method incorporates a significantly simple training process and does not require any handcrafted processing. We studied our network with real-world LDR images to confirm the feasibility in unknown data samples. 

Our contributions are as follows:

\begin{itemize}
    \item A two-stage deep network to reconstruct 16-bit HDR images from 8-bit LDR inputs.
    \item Comparison with state-of-the-art methods and outperform them in both objective and subjective measurement.
    \item Collection of an LDR image dataset and extensively study the proposed method's feasibility in real-world scenarios.
\end{itemize}

\section{Related Works}
LDR to HDR image reconstruction has been largely investigated in the last couple of years. The following subsection discusses some of the previous work on this topic, and for simplicity of the presentation, we categories those methods into learning and non-learning based methods. 

\subsection{Non-learning Based Methods}
Inverse tone-mapping \cite{banterle2006inverse}, additionally known as Expansion operators(EOs), broadly used for LDR to HDR image reconstruction,  has been studied for the last couple of decades. Nevertheless, this technique's difficulty persists as it lacks to produce details of the missing portion of the image. Hereabouts, concerning single image HDR reconstruction, we discuss some existing EOs techniques. EOs is commonly formulated mathematically as:

\begin{equation}
   L_e = f(L_d), where f : [0,255] \to R^+ 
\end{equation}

Here, $L_e$ indicates the produced HDR content from LDR inputs, which is denoted as $L_d$. f(.) indicates the expansion function, which takes LDR content as input.

Inverse tone mapping, along with global operators, mainly used in the early time of solving this LDR to HDR conversion problem. Landis \cite{landis2002production}, one of the earliest to solve this problem, used a linear function to all the images’ pixels. A gamma function has been used in Bist et al. \cite{bist2017tone} paper, where the gamma curve is defined with the help of the human visual system’s characteristics. Maisa et al. \cite{masia2017dynamic} proposed a global method that expands the content based on image properties determined by an image key. All the above methods are categorized as the global method \cite{banterle2017advanced}.

An analytical method coupled with an expand map is typically applied in the local method to expand LDR content to HDR. A median-cut \cite{debevec2008median} method was used in Banterle et al. \cite{banterle2006inverse} paper to find the areas with high luminance. Later they generated an expand map using an inverse operator to extend the luminance range in the high luminance areas. To maintain the contrast, Rempel et al. \cite{rempel2007ldr2hdr} further used an expand map calculated by a gaussian filter and an edge-stopping function.

Some other methods were proposed to tackle this issue where user interaction was added in most of them. Didyk et al. \cite{didyk2008enhancement} used a semiautomatic classifier to detect the high luminance and other saturated areas. Wang et al. \cite{wang2007high} proposed an impainting-based method where textures are recovered by transferring details from the user’s specific selected region. However, these above techniques solve LDR to HDR conversion problem and produce satisfactory outcomes only when well-behaved inputs are provided.

\subsection{Learning Based Methods}

Learning-based image to image translation like image enhancement showed great promises in the past decade. Considering their success in different domains of image manipulation, recent LDR to HDR  studies have incorporated deep learning in their respective solutions. In recent work, Endo et al. \cite{endoSA2017} propose an auto-encoder to generate HDR images from multi-exposure LDR images. Lee et al.\cite{lee2018deep} sequentially bracketed LDR exposures and utilized a CNN to reconstruct an HDR image. Later, Lee et al. \cite{lee2018deepGan} proposed a recursive conditional generative adversarial network (GAN) \cite{goodfellow2014generative} and combined an L1-norm to reconstruct the HDR images. Yu-Lun et al. \cite{liu2020single} intended to learn reverse camera pipeline for HDR reconstruction from a single input. Notably, all of these deep methods incorporate complicated training manoeuvre and handcrafted pre/post-processing operations. 

Apart from these approaches, a few novel methods propose to learn LDR to HDR directly through a single-stage deep network. For example, Eilertsen et al. \cite{eilertsen2017hdr} propose to utilize a U-Net architecture \cite{ronneberger2015u} to estimate the over-exposed region of an image and combines it with under-exposed pixels of the LDR inputs. In another way, Marnerides et al. \cite{marnerides2018expandnet} proposed a multi-branch CNN to extract features from the input LDR and fuse the output of each branch to expand the bit values of LDR images. Similarly, Zeeshan et al. \cite{khan2019fhdr} proposed a recurrent neural network to learn single-shot LDR to HDR from training pairs. The existing straightforward deep networks learn CRF and bit-expansion with a single-stage network, which can easily misinterpret the reconstruction network to produce visual artefacts. 

Unlike the existing works, the proposed method does not include any additional pre/post-processing operation. Our proposed method directly learns an 8-bit LDR to 16-bit HDR mapping with a novel deep network.

\section{Method}
The proposed method aims to recover 16-bit HDR images from single-shot LDR inputs. This section describes the process of network design, optimization, and implementation strategies in detail.

\begin{figure*}[!htb]
\centering
\includegraphics[width=\textwidth,keepaspectratio]{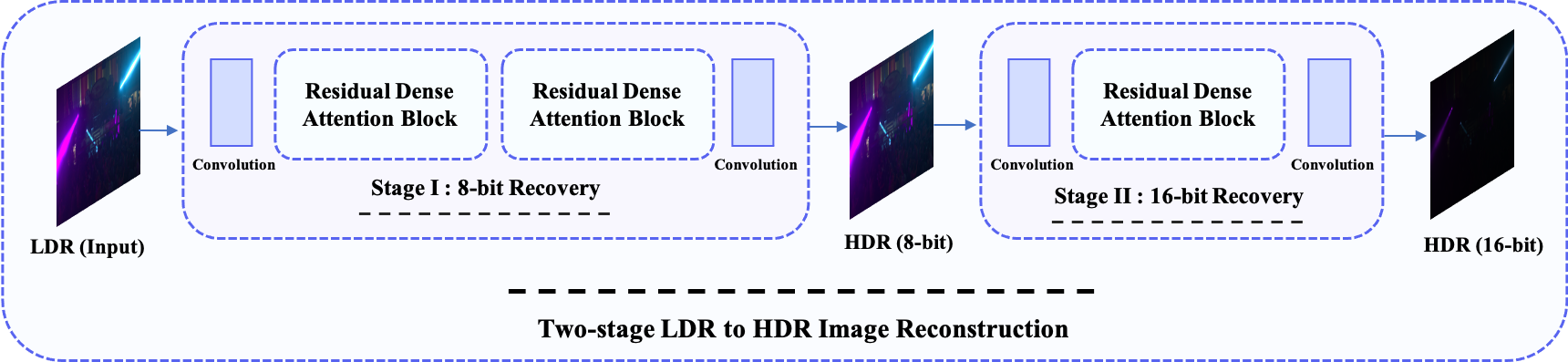}
\caption{  Overview of the proposed method. The proposed method comprises a two-stage deep network. Stage-I aims to perform image enhancement task such as denoising, exposure correction, etc. Stage-II of the proposed method intends to perform tone mapping and bit-expansion.  }
\label{overview}
\end{figure*}

\subsection{Network Design}
We consider the single-shot LDR to HDR formation as an image to image translation task. Therefore, the proposed deep network aims to recover 16-bit HDR images as $\mathrm{F}: I_L \to  I_H$. Where mapping function ($\mathrm{F}$) learns to generate a 16-bit image ($I_H$) from an 8-bit LDR image ($I_L$) comprehensively from a convex set of training samples. Fig. \ref{overview} illustrates the overview of the proposed method.

As Fig. \ref{overview} depicts, the proposed method comprises a two-stage deep network to map an input LDR input to an HDR image. The stages of the proposed deep method aim to perform as follows:

\begin{itemize}
    \item \textbf{Stage I:} Learns basic operation like exposure correction, denoising, contrast correction, gamma correction, etc. 
    \item \textbf{Stage II:} Learns tone mapping, bit-expansion, and recover 16-bit HDR images from the output of stage-I.
\end{itemize}

\textbf{Stage-I design.}  Typically, the LDR images illustrate numerous shortcomings like over/under exposure, over/desaturation, sensor noises, etc. Stage-I of the proposed method aims to perform such image enhancement tasks before reconstructing the HDR images.  Here, the network maps the input LDR input ($I_L$) as $I_{H^{\prime}} \in [0,M]^{H \times W \times  3}$. Here, $H$ and $W$ represent the height and width of $I_{H^{\prime}}$. The maximum value of $M$ can be perceived as $M=255$. However, we normalized the value of $M$ by dividing 255 to accelerate the training process. We design our stage-I as a stacked CNN and comprises a single convolutional operations (i.e., as input and output layer) with multiple Residual Dense Attention blocks (RDAB). To perceive a deeper architecture, we emphatically selected the frequency of RDAB in stage-I as RDAB  as $n=2$.  

\textbf{Stage-II design.} Stage-II of the proposed method aims to reconstruct the final 16-bit HDR images by learning tone mapping and bit expansion. Here, it takes the output of the stage-I  $I_{H^{\prime}}$ as input and maps it as $I_H \in [0, K]^{H \times W \times 3}$. It is noteworthy that the output range of $I_H$ has been stored in a 16-bit image format. Therefore, the maximum value of $K$ can be $K=65535$. Apart from that, this stage shares a similar network architecture as its predecessor. However, due to reduce the trainable parameter, we set the frequency of RDAB in stage-II as $n=1$.

\begin{figure}[!htb]
\centering
\includegraphics[width=\textwidth,keepaspectratio]{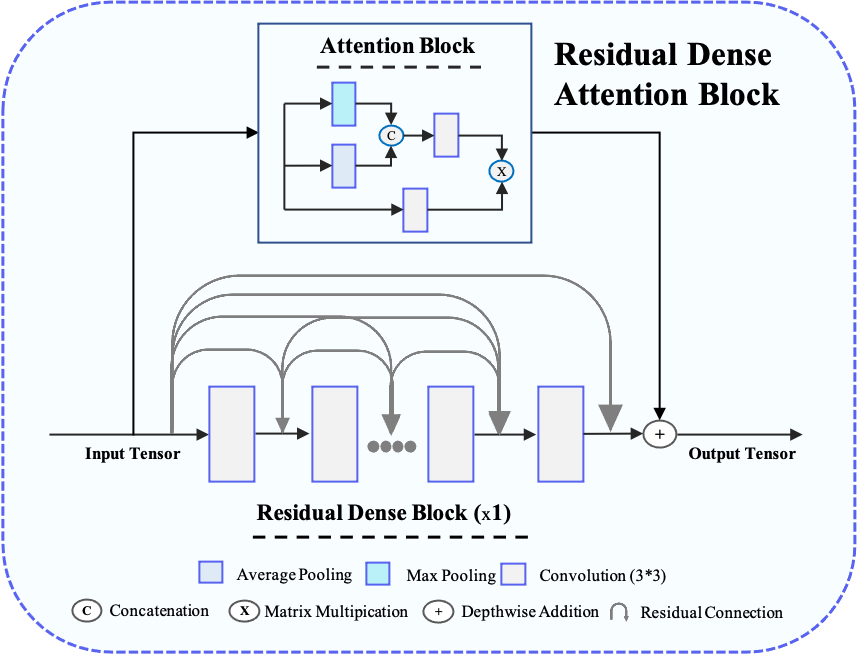}
\caption{ The residual dense attention block comprises a residual dense block and a spatial attention block. The stages of the proposed method leverage this residual dense attention block to accelerate the learning process. }
\label{rdab}
\end{figure}

\textbf{Residual Dense Attention Block.}
To accelerate our learning process, we develop a novel block combining a residual dense block \cite{zhang2018residual} and a spatial attention module \cite{woo2018cbam}, as shown in Fig. \ref{rdab}. Notably, the spatial attention modules in the newly developed RDAB  allowed us to leverage spatial attention along with residual feature propagation to mitigate visual artefacts. For a given input $X$, an RDAB aims to output the feature map ($X^{\prime}$) as:  
\begin{equation}
X = R(X) + {S}(X)
\label{dabEQ}
\end{equation}

$R(\cdot)$ and $S(\cdot)$ present the function of residual dense attention block and spatial attention block. We added the output of $S(\cdot)$ along with $R(\cdot)$ to learn a long-distance feature inter-dependency while performing HDR mapping. 

\subsection{Optimization}
The stages of the proposed method have been optimized with dedicated loss functions. Based on their dedicated role, we set the objective functions to maximize the performance.

\textbf{Stage-I optimization.} Typically, the deep networks have to employ a reconstruction loss to minimize the objective loss \cite{sharif2020learning}. This study utilizes an L1-norm as a base reconstruction loss \cite{schwartz2018deepisp}, which can be derived as follows:

\begin{equation}
 \mathcal{L}_{\mathit{R1}} = \parallel I_{G8}-I_{H^{\prime}}\parallel_1
\end{equation}

Here,  $I_{H^{\prime}}$ and $I_{G8}$ present the output obtained from stage-I and reference 8-bit image.

Due to the presence of extensive sensor noises in the LDR inputs, the generated images through the deep model can suffer from structural distortion. To avoid such unexpected structural degradation, we leveraged an SSIM loss \cite{schwartz2018deepisp, zhao2016loss} as structure loss and derived as follow:
\begin{equation}
 \mathcal{L}_{\mathit{S}} = SSIM \Big( I_{G8} , I_{H^{\prime}}\Big)
\end{equation}
 
We used a multi-scale variant of SSIM-loss during training.

Apart from the L1 and SSIM loss, we utilized a GAN based loss in this study. Here, the GAN-based loss aims to improve the texture in the reconstructed images \cite{ignatov2017dslr, wang2018esrgan} and derived as follows:

\begin{equation}
 \mathcal{L}_{\mathit{G}}= - \sum_{t} \log D(I_{H^{\prime}}, I_{G8})
\end{equation}
 
The total loss of the stage-I can be derived as :  

\begin{equation}
 \mathcal{L}_{\mathit{S1}}= \mathcal{L}_{\mathit{R1}} + \mathcal{L}_{\mathit{S}} +  1e-4.\mathcal{L}_{\mathit{G}}
\end{equation}

\textbf{Stage-II optimization.}
Similar to stage-I, we develop another dedicated loss function to maximize the performance of stage II. Here, the objective reconstruction loss of stage-II has obtained as follow:

\begin{equation}
 \mathcal{L}_{\mathit{R2}} = \parallel I_{G}-I_H\parallel_1
\end{equation}

Here,   $I_{H}$ and $I_{G}$ generated 16-bit HDR image and corresponding reference 16-bit image.

We combined a perceptual colour loss (PCL) \cite{sharif2021BJDD} along with the L1 loss to optimize stage-II. Here, the PCL aims to guide the network to avoid any colour degradation while mapping the given 8-bit images into a 16-bit HDR image \cite{sharif2021BJDD}. The PCL can be derived as follows:

\begin{equation}
 \mathcal{L}_{\mathit{C}} = \Delta{E} \Big( I_{G}, I_H \Big)
\end{equation}

Here, $\Delta{E}$ represents the CIEDE2000 colour difference between generated image and the reference image \cite{luo2001development}.

The total loss of stage-II can be summarized as follows:
\begin{equation}
 \mathcal{L}_{\mathit{S2}}= \mathcal{L}_{\mathit{R2}} + \mathcal{L}_{\mathit{C}} 
\end{equation}

\subsection{Implementation Details}
Both stages of the proposed method comprise a similar network architecture. The input layer of both stages aims to map an arbitrary image with a dimension of ${H \times W \times 3}$ into a feature map  $Z = {H \times W \times 64}$, where $H$ and $W$ represent the height and width of the input image. Contrarily, each network's output layer generates images as $I_{R^\ast} ={H \times W \times 3}$. The convolution operations of stage-I and stage-II comprises a kernel= $3 \times 3$, a stride=1,  padding=1, and activated by a ReLU activation. 

Apart from the stage-I and stage-II networks, the proposed method also utilizes a discriminator for estimating the adversarial loss. Here, we adopted a well-established variant of the generative adversarial network (GAN) known as conditional GAN (cGAN) \cite{mirza2014conditional, liu2020importance} to obtain a stable training phase. Our discriminator's goal has set to maximize $\mathbb{E}_{X,Y} \big[\log D\big(X,Y\big) \big]$. The network comprises eight consecutive convolutional layers with a  kernel size of $3 \times 3$ and activated with a swish function. The feature depth of these convolutional layers has started from 64 channels. In every $(2n-1)^{th}$ layer, the architecture expands its feature depth and reduce the spatial dimension by a factor of 2. The final output of the discriminator obtained with another convolution operation comprising a kernel = $1 \times 1$ and activated by a sigmoid function.

\section{Experiments and results}

We perform dense experiments to study the feasibility of the proposed study in a different scenario. This section details the results obtained from the experiments for LDR to HDR reconstruction.

\subsection{Setup}

\begin{figure}%

\captionsetup{width=\textwidth}
\captionsetup[subfigure]{}
\centering
 \begin{minipage}{\textwidth}
    \begin{subfigure}{\textwidth}
      \centering
        \includegraphics[width=\textwidth]{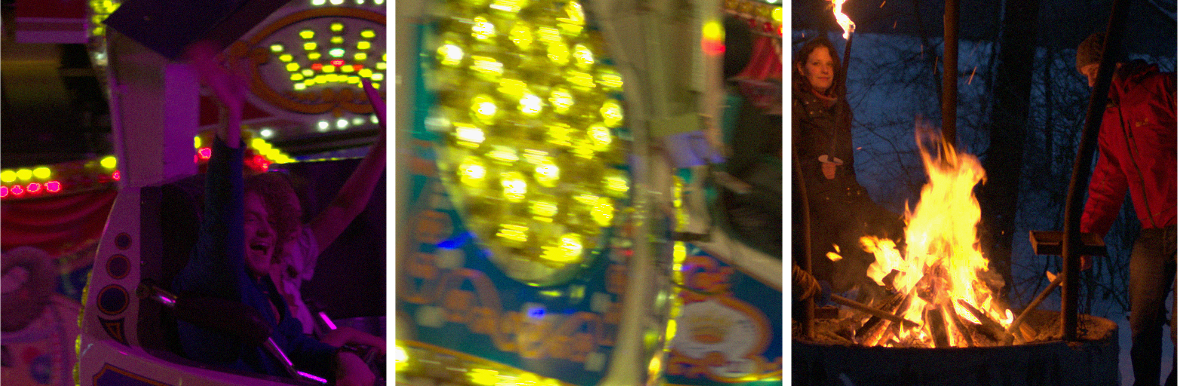}
    \end{subfigure}
    \begin{subfigure}{\textwidth}
      \centering
      \includegraphics[width=\textwidth]{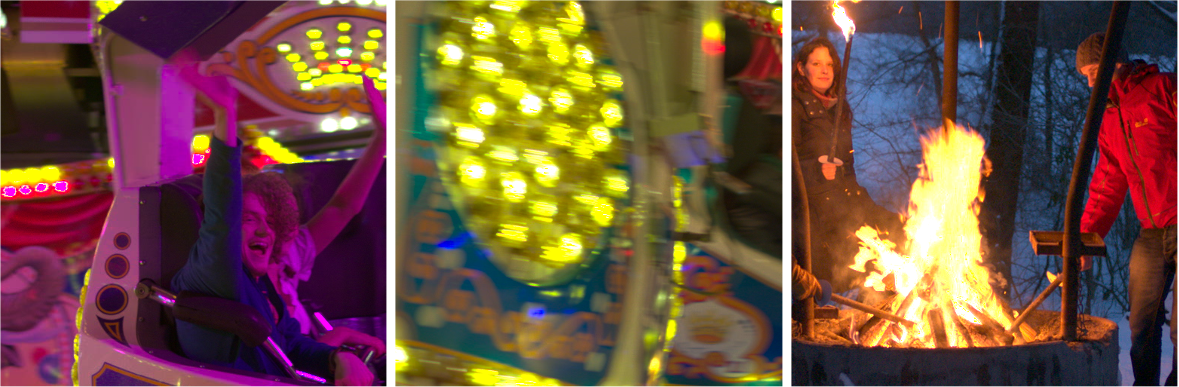}
    \end{subfigure}
    \begin{subfigure}{\textwidth}
      \centering
      \includegraphics[width=\textwidth]{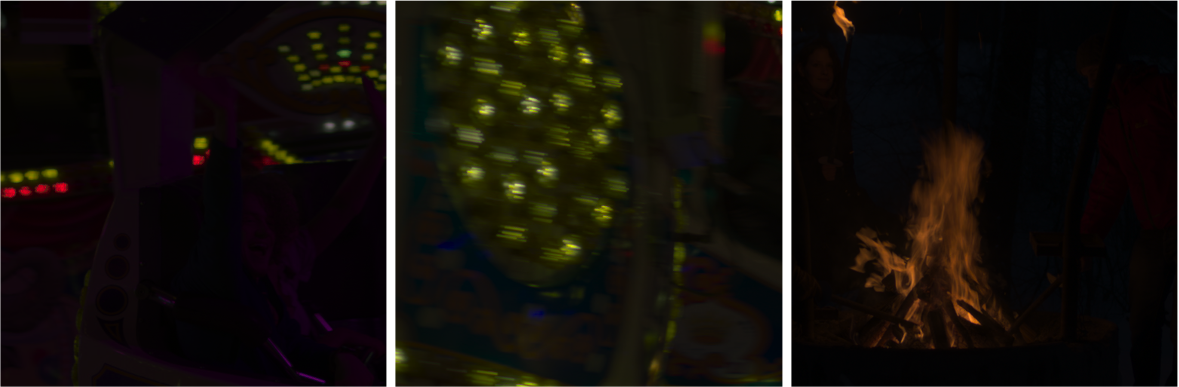}
    \end{subfigure}
     
  \end{minipage}

\caption{ Example of image patches used for training. Top row: LDR image (Input), middle row: reference image (8-bit), bottom row: reference image (16-bit). }
 
\label{impPatch}
\end{figure}

We studied our method with images from the HdM HDR dataset \cite{singleHDRChallange,hdr2021ntire}. The used dataset comprised a set of 1289 scenes (i.e., long, medium, short exposure LDR images, and 16-bit HDR ground-truth) captured with two Alexa Arri cameras. For this study, we used 1,000 image sets for training and the rest for the testing.  We extracted a total of 7,551  image patches and made image sets for exploiting supervised training. Each patch set comprised randomly extracted images patches of LDR input, 16-bit and 8-bit ground truth images. It is worth noting, we obtained the 8-bit reference images by clipping and normalizing the 16-bit ground truth images. Fig. \ref{impPatch} depicts the sample image patches that we used extracted from the HdM HDR dataset, which we used for training only. Apart from that, we evaluated our method with higher resolution images in the later stages.

The proposed solution is implemented with the PyTorch framework \cite{pytorch}. Additionally, the networks were optimized with an  Adam optimizer \cite{kingma2014adam} , where the hyperparameters were tuned as $\beta_1 = 0.9$, $\beta_2 = 0.99$, and learning rate = 5e-4.  We trained our model for 25 epochs with a constant batch size of 8. It took around 24 hours to converge our model. We conducted our experiments on a machine comprises of an AMD Ryzen 3200G central processing unit (CPU) clocked at 3.6 GHz, a random-access memory of 16 GB, and An Nvidia Geforce GTX 1060 (6GB) graphical processing unit (GPU).

\subsection{Comparison with state-of-the-art methods}

We compared our methods with three different state-of-the-art single-shot LDR to HDR works: i) HDRCNN \cite{eilertsen2017hdr}, ii) ExpandNet \cite{marnerides2018expandnet}, and iii) FHDR \cite{khan2019fhdr}. It is worth noting, none of these methods has been specially designed for generating 16-bit HDR images, as we aim to learn in this study. However, to keep the evaluation process as fair as possible, we studied each state-of-the-art model with the same dataset we used to investigate our proposed method. We trained these single-shot HDR reconstruction networks with pairs of reference 16-bit images and input LDR images. Also, each method was studied with their suggested hyperparameters until they converge with the given data samples. We evaluated each deep method with the same testing samples and summarized the performance with peak-signal-to-noise-ratio (PSNR) and $\mu$-PSNR metrics \cite{hdr2021ntire}. Here, we compute the $\mu$-PSNR as per the suggestions of \cite{hdr2021ntire} and employed a compression factor $\mu = 5000$, normalizing percentile = $99$, and a $\tanh$ function for maintaining the $\left[0,1\right]$ range.

\begin{table}[!htb]
\begin{tabular}{lll}
\hline
\textbf{Method} & \textbf{PSNR}  & \textbf{$\mu$-PSNR} \\ \hline
HDRCNN \cite{eilertsen2017hdr}         & 31.46          & 24.30            \\
ExpandNet \cite{marnerides2018expandnet}       & 32.77          & 29.84            \\
FHDR \cite{khan2019fhdr}           & 33.33          & 31.15            \\
\textbf{Ours}   & \textbf{34.29} & \textbf{32.66}  \\\hline
\end{tabular}
\caption{Quantitative comparison between the proposed method and existing learning-based single-shot LDR to HDR methods. The proposed method outperforms the state-of-the-art methods in both evaluation metrics.}
\label{compTable}
\end{table}

\begin{figure*}[!htb]%
\centering
\captionsetup[subfigure]{labelformat=empty}
\begin{minipage}{.46\textwidth}
    \begin{subfigure}{\textwidth}
      \centering
     \includegraphics[width=\textwidth]{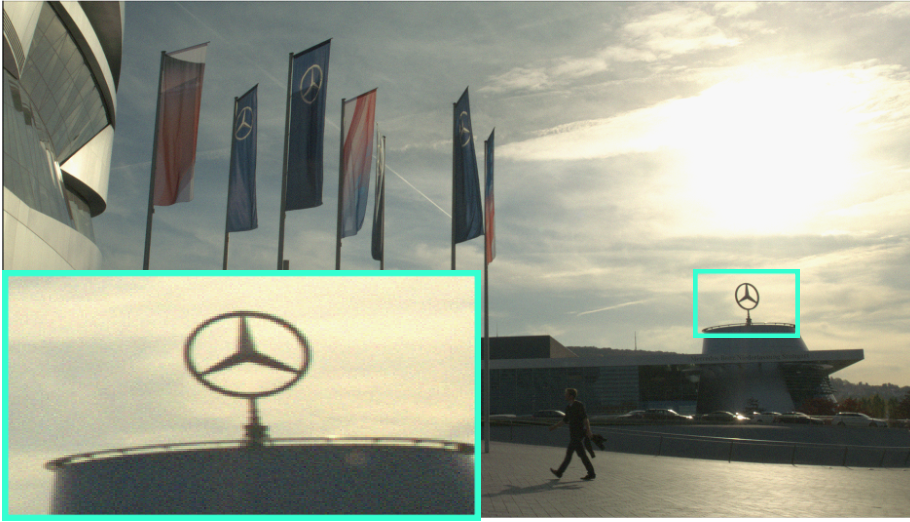}
     \caption{LDR (input)}
    \end{subfigure}
  \end{minipage}
  \begin{minipage}{.22\textwidth}
    \begin{subfigure}{\textwidth}
      \centering
        \includegraphics[width=3.8cm]{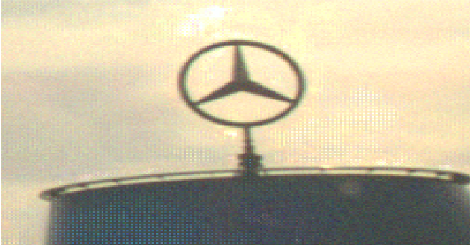}
        \caption{HDRCNN \cite{eilertsen2017hdr}}
    \end{subfigure}
    \begin{subfigure}{\textwidth}
      \centering
      \includegraphics[width=3.8cm]{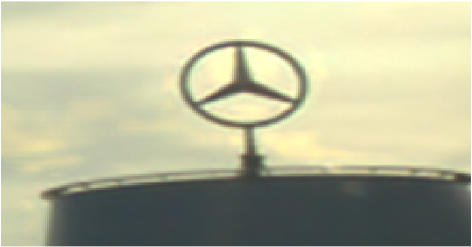}
      \caption{FHDR \cite{khan2019fhdr}}
    \end{subfigure}
  \end{minipage}
  \begin{minipage}{.22\textwidth}
    \begin{subfigure}{\textwidth}
      \centering
        \includegraphics[width=3.8cm]{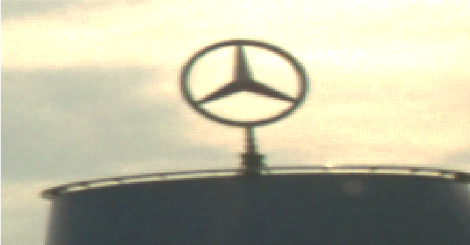}
        \caption{ExpandNet \cite{marnerides2018expandnet}}
    \end{subfigure}
    \begin{subfigure}{\textwidth}
      \centering
      \includegraphics[width=3.8cm]{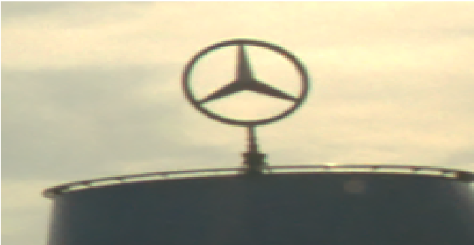}
      \caption{\textbf{Our} }
    \end{subfigure}
  \end{minipage}

  \caption{ Quantitative comparison between proposed method and existing learning-based single-shot LDR to HDR methods. }
\label{compSOTA}
\end{figure*}

 \textbf{Quantitative evaluation.} Table \ref{compTable} illustrates the quantitative comparisons between the deep methods. It can be seen that our two-stage HDR reconstruction method outperforms the existing deep methods in both evaluation metrics with a marginal score. It scores 34.29 dB in PSNR and 32.66 dB in $\mu$-PSNR metrics, which is almost 3dB and 8dB higher in PSNR and $\mu$-PSNR metrics than the lowest-performing deep network (i.e., HDRCNN \cite{eilertsen2017hdr}).  It is worth noting the HDRCNN \cite{eilertsen2017hdr} model leverage a VGG-16 backbone \cite{simonyan2014very} in its architecture. Typically, such pre-trained VGG-16 backbone networks aim to enhance the details while performing any image to image translations task.  We found that the VGG-16 backbone of HDRCNN boosts the sensor noise of LDR inputs while detail enhancement. Also, the 16-bit expansion boosts up these noises further in final reconstruction and resist the HDRCNN to perform a satisfactory performance as its counterparts.

\textbf{Qualitative comparison.} Apart from the quantitative comparison, we perform a qualitative evaluation to perform the subjective measurement between the different single-shot LDR to HDR reconstruction methods. Fig. \ref{compSOTA} illustrates reconstructed HDR images obtained through the different deep models. We normalized and clipped the 16-bit HDR outputs for better visualization. The visual comparison grasps the consistency of quantitative comparison. Moreover, our two-stage deep method reconstructs cleaner HDR images with natural colour consistency. It maintains the details in the complicated overexposed regions comparing to its counterparts.  Overall, the proposed method can recover plausible HDR image from an LDR input without producing any visually disturbing artefacts.

\subsection{Ablation Study}

We studied the feasibility and the contribution of our two-stage method with sophisticated experiments. Specifically, we trained and evaluated our stages separately to verify the feasibility of a two-stage model for LDR to HDR reconstruction. Here, we used challenging single-shot LDR images from the HdM HDR dataset to perform the quantitative and qualitative evaluation.

\begin{table}[!htb]
\begin{tabular}{lll}
\hline
\textbf{Method}       & \textbf{PSNR}  & \textbf{$\mu$-PSNR} \\ \hline
Stage-I               & 31.71          & 18.43            \\
Stage-II              & 29.42          & 15.73            \\
\textbf{Stage I + II} & \textbf{34.29} & \textbf{32.66}  \\ \hline
\end{tabular}
\caption{Ablation study of the proposed method. We performed a quantitative evaluation with images from the HdM HDR dataset.}
\label{ablTable}
\end{table}

\textbf{Quantitative evaluation.} Table. \ref{ablTable} illustrates the performance of each stage of the proposed method on the HdM HDR dataset. Here, the PSNR and $\mu$-PSNR calculated over 289 image pairs. We arbitrarily selected an LDR image from the three exposure shoots and paired it with the ground truth image for performing the evaluation. The ablation study illustrates that each stage of the proposed method contributes to the final HDR reconstruction. The individual stages of the proposed method can not achieve the height in evaluation metrics as their two-stage variants. We observed a tendency of underfitting in one-stage variants due to the significantly lesser number of trainable parameters (please see sec. \ref{discussion} for detail).

\begin{figure*}%
\centering
\captionsetup[subfigure]{labelformat=empty}

  \begin{minipage}{.24\textwidth}
    \begin{subfigure}{\textwidth}
      \centering
        \includegraphics[width=4.2cm]{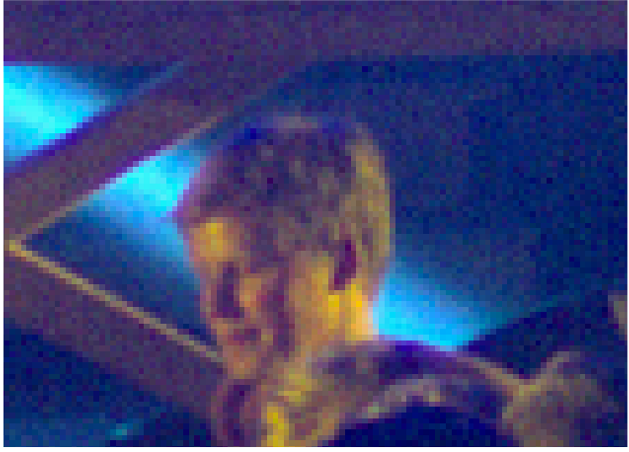}
        \caption{LDR (Input)}
    \end{subfigure}
  \end{minipage}
  \begin{minipage}{.24\textwidth}
    \begin{subfigure}{\textwidth}
      \centering
        \includegraphics[width=4.2cm]{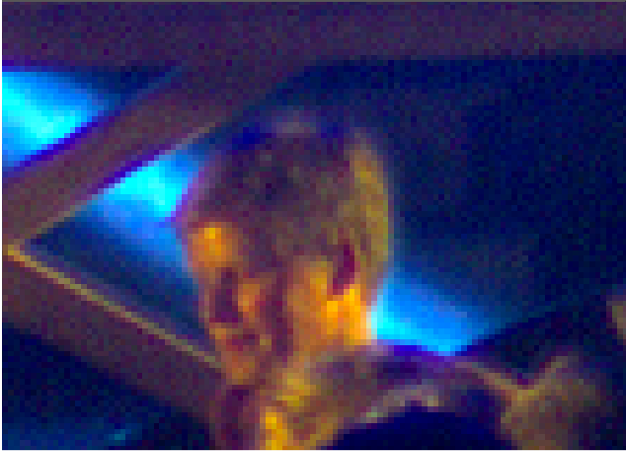}
        \caption{Stage-I (Visualized)}
    \end{subfigure}
  \end{minipage}
  \begin{minipage}{.24\textwidth}
    \begin{subfigure}{\textwidth}
      \centering
        \includegraphics[width=4.2cm]{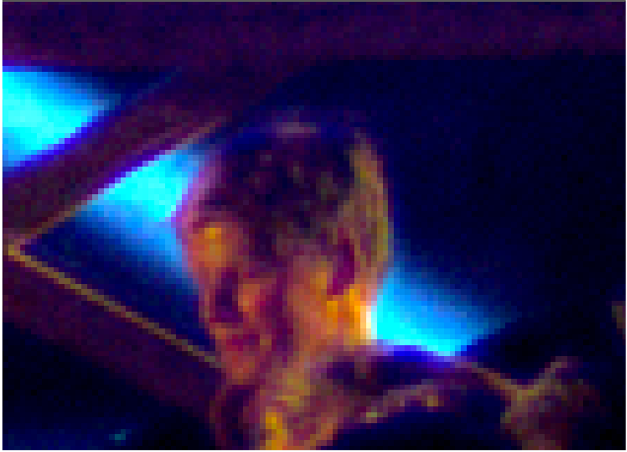}
        \caption{Stage-II (Visualized)}
    \end{subfigure}
  \end{minipage}
    \begin{minipage}{.24\textwidth}
    \begin{subfigure}{\textwidth}
      \centering
        \includegraphics[width=4.2cm]{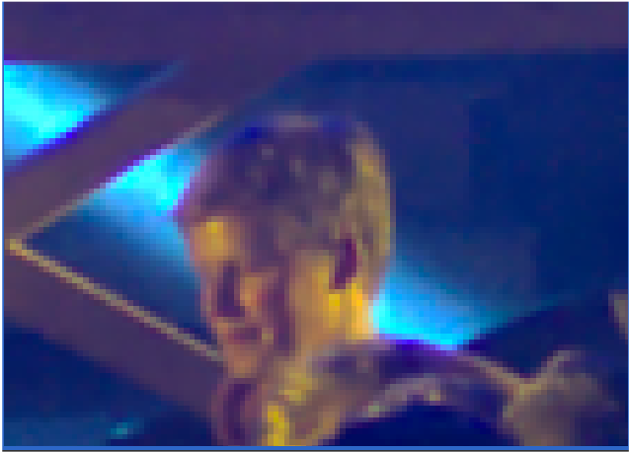}
        \caption{Stage-I + Stage-II (Visualized)}
    \end{subfigure}
  \end{minipage}

  \caption{ Qualitative evaluation on different variants of the proposed method. The proposed two-stage variants can reconstruct better HDR images than its one-stage variants. }
\label{ablVis}
\end{figure*}

\textbf{Qualitative evaluation.}  Fig. \ref{ablVis} illustrates the visual comparison between different variants of the proposed study. Results have been visualized by applying a normalizing factor on 16-bit HDR images. It can be visible that the proposed two-stage model can reconstruct visually cleaner and plausible images among all models.  Despite sharing similar network configurations, the single-stage networks struggle to reach the height of their two-stage variants. Particularly, estimating CRF, bit-expansion with image enhancement misinterpreted them to produce visual artefacts. 

\subsection{Method generalization}
The key motivation of our proposed works is to obtain satisfactory results on diverse LDR images. Therefore, we studied the feasibility of our proposed method with a substantial amount of LDR samples captured with different hardware. To obtain this, we collected an LDR dataset incorporating numerous camera hardware, including DSLR (i.e., Canon Rebel T3i) and smartphone cameras (i.e., Samsung Galaxy Note 8, Xiaomi MI A3, iPhone 6s, etc.). We collected a total of 52 LDR images using these devices. Depending on the hardware types (i.e., DSLR or smartphone), we capture images by applying the following strategies:
\begin{itemize}
    \item \textbf{DSLR:} To capture LDR images with DSLR, we mostly used auto exposure settings and captured a total of 25 LDR shots with such configuration. Notably, we choose stochastic lighting conditions like middy sun, low-light condition,  high-contrast lighting condition, and sunset as shooting environments. Which allowed us to cover the most challenging shooting environments from real-world environments.
    
    \item \textbf{Smartphone:} Smartphone photography has gained significant popularity over the last decades. Therefore, we included images capture with different smartphone cameras in our LDR dataset. Typically, due to the shortcoming of smaller sensor size \cite{ignatov2020replacing, ignatov2017dslr, sharif2021BJDD}, smartphone OEMs shipped their devices with the ability to produce HDR images. However, such default HDR settings do not fit well with our target applications. Thus, we used a third-party camera app known as Open Camera for capturing the LDR images with different smartphones. We disabled the HDR mode, including HDR contrast enhancement from the default settings of the application. Apart from that, we kept the exposure setting in auto mode and captured a total of 27 LDR images in tricky lighting conditions similar to the DSLR setup.
   
\end{itemize}

The collected images presented into a unified dataset and resampled into $2048 \times 1080 \times 3$ resolution. Later, we inference the resampled LDR images with our proposed method and summarized the results with a visual and blind-fold user study.

\begin{figure*}[!htb]
\centering
\captionsetup[subfigure]{labelformat=empty}

  \begin{minipage}{.32\textwidth}
    \begin{subfigure}{\textwidth}
      \centering
        \includegraphics[width=5.6cm]{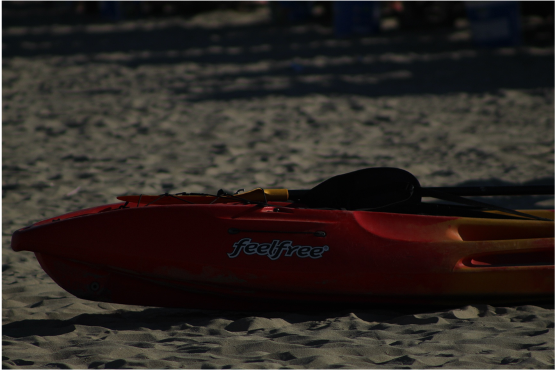}
    \end{subfigure}
    \begin{subfigure}{\textwidth}
      \centering
      \includegraphics[width=5.6cm]{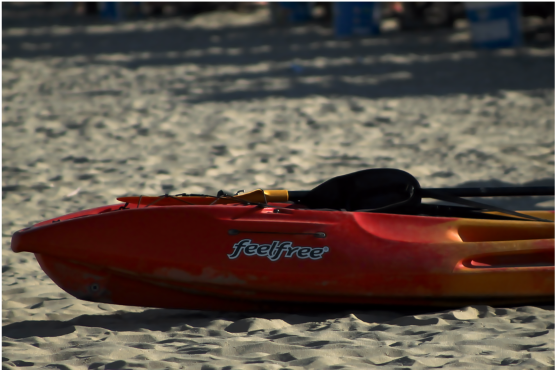}
    \end{subfigure}
  \end{minipage}
  \begin{minipage}{.32\textwidth}
    \begin{subfigure}{\textwidth}
      \centering
        \includegraphics[width=5.6cm]{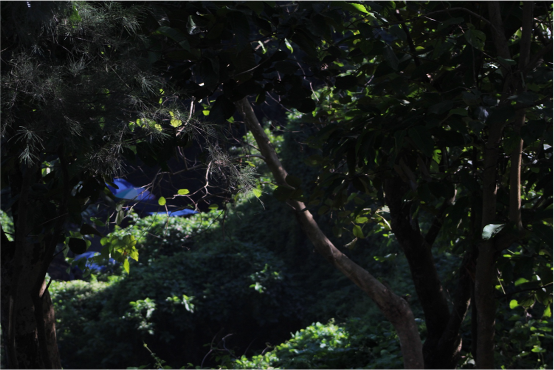}
    \end{subfigure}
    \begin{subfigure}{\textwidth}
      \centering
      \includegraphics[width=5.6cm]{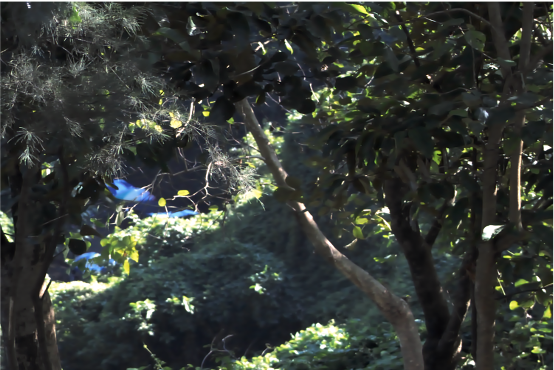}
    \end{subfigure}
  \end{minipage}
  \begin{minipage}{.32\textwidth}
    \begin{subfigure}{\textwidth}
      \centering
        \includegraphics[width=5.6cm]{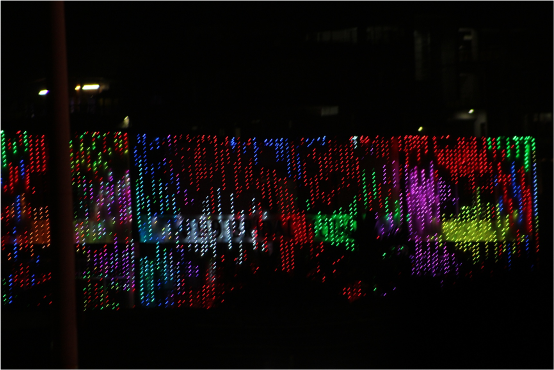}
    \end{subfigure}
    \begin{subfigure}{\textwidth}
      \centering
      \includegraphics[width=5.6cm]{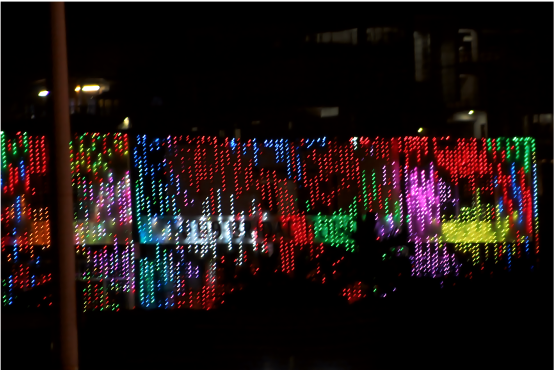}
    \end{subfigure}
  \end{minipage}

  \caption{ Real-world LDR to HDR reconstruction obtained by the proposed study. Top row: LDR images, bottom row: reconstructed HDR images (visualized) }
\label{genVis}
\end{figure*}

\textbf{Visual Results.} Fig. \ref{genVis} depicts the real-world LDR to HDR mapping obtained by the proposed method. Our method maps an 8-bit LDR image into the 16-bit HDR image. However, due to better visualisation, we clipped and normalised the 16-bit images into an 8-bit format. Despite the clipping process, it can be observable that the proposed method can handle LDR images captured with the diverse camera hardware without explicitly knowing the CRF and exposure settings. Also, the proposed method does not require any additional pre/post-processing operations. 

\textbf{User study.} A blind-fold user study has been performed to summarise the preferences of random users on our HDR reconstruction. We conducted the evaluation process on 50 users, where users were age between $\left[10,60\right]$. We showed ten random image pairs to each participating user, where image pairs comprised an LDR input and our reconstructed HDR image (clipped and normalized). Later, we allowed the users to pick an image from the image pairs as their personal preference.  We conducted the user study anonymously, and the information related to the evaluation process remained secret to the participating users. We summarized the unbiased user opinion with a mean opinion score (MOS). Table. \ref{mos} shows the MOS obtained by conducting our user preference study. The proposed single-shot HDR reconstruction method outperforms LDR images in blind-fold testing by a substantial margin. Also, the user study reveals the feasibility of the proposed two-stage deep network in HDR image reconstruction for consumer-grade camera systems.

\begin{table}[!htb]
\centering
\begin{tabular}{lll}
\hline
\textbf{Method} & \textbf{MOS $\uparrow$} \\ \hline
 LDR (Input)                            &       1.2  \\\hline
 \textbf{HDR (Reconstructed)} &       \textbf{3.8}  \\\hline
\end{tabular}

\caption{A user study on LDR image(input) and HDR image(output). Higher MOS indicates better user preference.}
\label{mos}
\end{table}

\subsection{Discussion}
\label{discussion}
The proposed method is developed to participate in NTIRE 2021 High Dynamic Range Challenge (Track 1 Single Frame) \cite{singleHDRChallange,hdr2021ntire}. In the final competition, we secured the top five position with our fully convolutional solution. Our method scored 30.99 and 32.84 respectively in PSNR and $\mu$-PSNR metrics \cite{hdr2021ntire}. 

The proposed method comprises 834,476 trainable parameters (555,655 for stage-I and 278,821 for stage-II). Despite train with images patches, our model can be inference with any dimensioned images. Our model takes around 1.10 seconds to successfully inference an image dimensioned of $1900 \times 1060 \times 3$. As the proposed method doesn't require any pre/post-processing, the inference times are meant to remain contents with the same hardware settings. Subsequently, the simplicity of the proposed method made the solution convenient for real-world deployment.

\section{Conclusion}

This study proposed a two-stage learning-based method for single-shot LDR to HDR mapping without explicitly calculating camera hardware related information. Here, stage-I of the proposed method learns to perform the basic image manipulation techniques like exposure correction, denoising, brightness correction comprehensively. Additionally, stage-II focuses on tone mapping and bit-expansion to output 16-bit HDR images. We evaluated and compared our proposed approach with the state-of-the-art single-shot HDR reconstruction methods. Both qualitative and quantitative comparison evident that the proposed method can outperform the existing deep methods with a substantial margin. Apart from that, we also collected a set of LDR images captured with the different camera hardware. The study with our newly collected dataset reveals that the proposed method can handle the real-world LDR samples without producing any visual artefacts. It has planned to extend the proposed method for multi-shot HDR reconstruction in a future study. 

\section*{Acknowledgments}
This work was supported by the Sejong University Faculty Research Fund.

{\small
\bibliographystyle{ieee_fullname}
}

\end{document}